\documentclass{article}

\usepackage[final,nonatbib]{neurips_2022}

\usepackage[utf8]{inputenc} 
\usepackage[T1]{fontenc}    
\usepackage{url}            
\usepackage{booktabs}       
\usepackage{amsfonts}       
\usepackage{nicefrac}       
\usepackage{microtype}      
\usepackage{xcolor}         

\usepackage{graphicx}

\usepackage{nccmath}  
\usepackage{bbm}
\usepackage{amsmath}
\usepackage{amssymb}
\usepackage{mathtools}
\usepackage{amsthm}

\usepackage[flushleft]{threeparttable} 
\usepackage{multirow}

\usepackage{wrapfig}

\DeclareMathOperator*{\argmin}{arg\,min}

\usepackage{pifont}
\newcommand\mynuma[1]{\ifcase#1 \or \ding{172}\or \ding{173}\or
  \ding{174}\or \ding{175}\or \ding{176}\or \ding{177}%
  \or \ding{178}\or \ding{179}\or \ding{180}\or \ding{181}\else *\fi\relax}

\newcommand\mynumb[1]{\ifcase#1 \or \ding{182}\or \ding{183}\or
  \ding{184}\or \ding{185}\or \ding{186}\or \ding{187}%
  \or \ding{188}\or \ding{189}\or \ding{190}\or \ding{191}\else *\fi\relax}

\newcommand{\METHOD}{S$^3$-Router}

\title{Losses Can Be Blessings: Routing Self-Supervised Speech Representations Towards Efficient \\ Multilingual and Multitask Speech Processing}

\author{
  Yonggan Fu\textsuperscript{1}, Yang Zhang\textsuperscript{2}, Kaizhi Qian\textsuperscript{2}, Zhifan Ye\textsuperscript{3}, Zhongzhi Yu\textsuperscript{1} \\
  \textbf{Cheng-I Lai\textsuperscript{4}, Yingyan (Celine) Lin\textsuperscript{1}}\\
  \textsuperscript{1}Georgia Institute of Technology, \textsuperscript{2}MIT-IBM Watson AI Lab, \textsuperscript{3}Rice University, \textsuperscript{4}MIT CSAIL \\
  \texttt{\{yfu314,zyu401,celine.lin\}@gatech.edu} \\ \texttt{\{yang.zhang2,kqian\}@ibm.com} \,\, \texttt{\{zy50\}@rice.edu}  \,\, \texttt{\{clai24\}@mit.edu}\\
}

\begin{document}

\maketitle

\begin{abstract}

Self-supervised learning (SSL) for rich speech representations has achieved empirical success in low-resource Automatic Speech Recognition (ASR) and other speech processing tasks, which can mitigate the necessity of a large amount of transcribed speech and thus has driven a growing demand 
for on-device ASR and other speech processing.
However, advanced speech SSL models have become increasingly large, which contradicts the limited on-device resources. This gap could be more severe in multilingual/multitask scenarios requiring simultaneously recognizing multiple languages or executing multiple speech processing tasks. Additionally, strongly overparameterized speech SSL models tend to suffer from overfitting when being finetuned on low-resource speech corpus.
This work aims to enhance the practical usage of speech SSL models towards a win-win in both enhanced efficiency and alleviated overfitting via our proposed \METHOD{} framework, which for the first time discovers that simply discarding no more than 10\% of model weights via only finetuning model \textit{connections} of speech SSL models can achieve better accuracy over standard weight finetuning on downstream speech processing tasks. More importantly, \METHOD{} can serve as an all-in-one technique to enable (1) a new finetuning scheme, (2) an efficient multilingual/multitask solution, (3) a state-of-the-art ASR pruning technique, and (4) a new tool to quantitatively analyze the learned speech representation. 
We believe \METHOD{} has provided a new perspective for practical deployment of speech SSL models. Our codes are available at: \url{https://github.com/GATECH-EIC/S3-Router}.

\end{abstract}

\section{Introduction}
\label{sec:intro}

Deep neural network (DNN) breakthroughs have tremendously advanced the field of Automatic Speech Recognition (ASR). However, one major driving force for powerful DNNs, i.e., the availability of a large amount of training data, is not always possible for ASR. This is because collecting large-scale transcriptions is costly, especially for low-resource spoken languages around the world, limiting the wide application of deep ASR models. Fortunately, recent advances in self-supervised learning (SSL) for rich speech representations~\cite{chi2020audio,ling2020decoar,oord2018representation, kharitonov2020data, schneider2019wav2vec, baevski2020wav2vec, conneau2020unsupervised, baevski2019effectiveness, hsu2021hubert} have achieved empirical success in low-resource ASR, where SSL models pretrained on raw audio data without transcriptions can be finetuned on low-resource transcribed speech to match the accuracy of their supervised counterparts.

However, there exists a dilemma between the trends of speech SSL models and the growing demand for speech processing applications on the edge. While advanced speech SSL models become increasingly larger to learn more generalizable features, it is highly desired to process the captured speech signals in real time on edge devices, which have limited resources and conflict with the prohibitive complexity of existing speech SSL models. Such an efficiency concern would be more severe in multilingual/multitask scenarios where simultaneously recognizing multiple languages or executing multiple speech processing tasks is required: if one separate model is finetuned for each target language/task, the storage and computational cost will be significantly increased, thus prohibiting the practical deployment of existing speech SSL models. 
Additionally, strongly overparameterized speech SSL models tend to suffer from overfitting when being finetuned on a low-resource speech corpus~\cite{hou2021exploiting,cai2014convolutional,cai2014stochastic,chan2015deep,miao2013improving}, limiting the achievable accuracy improvement brought about by more parameters and thus the achievable performance-efficiency trade-off.

In this work, we aim to facilitate the practical usage of speech SSL models towards a win-win in enhanced efficiency and alleviated overfitting for boosting task accuracy under low-resource settings. 
\underline{Excitingly}, we develop a framework, dubbed \textbf{S}elf-\textbf{S}upervised \textbf{S}peech Representation \textbf{Router} (\METHOD{}), that can serve as an all-in-one technique to tackle the aforementioned challenges and largely enhance the practical usage of speech SSL models, contributing \underline{(1)} a new finetuning scheme for downstream speech processing, i.e., finetuning the connections of the model structure via learning a binary mask on top of pretrained model weights, which notably alleviates model overfitting and thus improves the achievable accuracy over standard weight finetuning methods under a low-resource setting; \underline{(2)} a multilingual/multitask technique via learning language-/task-specific binary masks on top of shared model weights inherited from SSL pretraining; \underline{(3)} a competitive ASR pruning technique to trim down the complexity of speech SSL models while maintaining task accuracy; and \underline{(4)} a new tool to quantitatively analyze what is encoded in speech SSL models thanks to the learned masks' binary nature on top of shared model weights.
We summarize our contributions below:

\begin{itemize}
    \item We propose a framework dubbed \METHOD{}, which offers an alternative to the mainstream finetuning of model \textit{weights} that finetunes the \textit{structure} of speech SSL models, by learning a binary mask on top of the pretrained model weights and integrating it with a novel mask initialization strategy customized for the pretrain-finetune paradigm;
       \vspace{-0.1em}
    \item We are the first to discover that discarding no more than 10\% of weights \textit{without} finetuning pretrained model weights can achieve better task performance as compared to the mainstream method of weight finetuning on downstream speech processing tasks. Notably, our method can scale well to even larger models;
       \vspace{-0.1em}
    \item We extend our \METHOD{} framework for enhanced deployment efficiency, contributing \underline{(1)} a novel multilingual/multitask solution, and \underline{(2)} a competitive pruning technique achieving better performance-efficiency trade-offs than state-of-the-art (SOTA) ASR pruning techniques;
    \vspace{-1.2em}
    \item We demonstrate the capability of \METHOD{} as a tool to quantitatively understand what is encoded in speech SSL models thanks to the binary nature of the learned masks.

\end{itemize}

\vspace{-0.6em}

\METHOD{} has opened up a new perspective for empowering efficient multilingual/multitask speech processing and enhancing our understanding about what is encoded in speech SSL models.

\section{Related Work}
\label{sec:related_work}

\textbf{Automatic speech recognition.}
Early ASR systems~\cite{sha2006large,mao2019revisiting,adams2019survey,tang2009hybrid,bridle1990alpha} were mainly based on the combinations of hidden Markov models (HMM) with Gaussian mixture models or DNNs, and often contain multiple modules (e.g., an acoustic model, a language model, and a lexicon model) trained separately. Recent works process raw audio sequences end-to-end, including CTC~\cite{graves2006connectionist}-based models~\cite{hannun2014deep, amodei2016deep, graves2014towards, miao2015eesen, eyben2009speech}, recurrent neural network(RNN)-transducers~\cite{graves2012sequence, graves2013speech, rao2017exploring, dong2018extending}, sequence-to-sequence models~\cite{chorowski2015attention, bahdanau2016end, chan2016listen, zhang2017very, chiu2018state, prabhavalkar2018minimum}, and transformer-based
models~\cite{gulati2020conformer, dong2018speech, wang2020transformer}. Specifically, transformer-based models have been widely adopted as speech SSL models~\cite{baevski2020wav2vec, hsu2021hubert, conneau2020unsupervised}.

\textbf{Self-supervised learning for speech representation.}
Considering the high cost of collecting large-scale transcriptions, learning rich speech representations via SSL has become crucial and promising for empowering low-resource ASR.
Early works~\cite{hsu2017learning,chorowski2019unsupervised,van2017neural,hsu2017unsupervised,ebbers2017hidden,glarner2018full,khurana2019factorial,khurana2020convolutional} build generative models for speech with latent variables. Recently, prediction-based SSL methods have become increasingly popular, where the models are trained to reconstruct the contents of unseen frames~\cite{chung2019unsupervised, chung2020generative, chung2020improved, ling2020deep, wang2020unsupervised, liu2020mockingjay, chi2020audio, ling2020decoar} or contrast the features of masked frames with those of randomly sampled ones~\cite{oord2018representation, kharitonov2020data, schneider2019wav2vec, baevski2020wav2vec, conneau2020unsupervised, baevski2019effectiveness, hsu2021hubert}. 
We refer the readers to a recent survey~\cite{liu2022audio} for more details. Among prior arts,~\cite{zhao2020masking} is a pioneering work relevant to our method, and adopts masking as an alternative to weight finetuning on pretrained language models for natural language processing (NLP). Nevertheless, our \METHOD{} is non-trivially different from~\cite{zhao2020masking} in that \underline{(1)} we target SSL models in the speech domain and the learned speech representation is required to be generalizable across different spoken languages under a cross-lingual transfer setting, where the speech SSL models suffer from a higher risk of overfitting on downstream low-resource tasks as compared to NLP, making our findings non-trivial contributions;
\underline{(2)} \METHOD{} features a win-win in both efficiency and accuracy thanks to
our proposed mask initialization strategy, which plays a crucial role in maintaining task accuracy under a high sparsity and enables the extension of \METHOD{} towards a competitive pruning technique with SOTA accuracy; and (3) we further develop \METHOD{} for multilingual/multitask speech processing and as a simple yet effective tool for analyzing language-wise similarities.

\textbf{Multitask learning for speech processing.}
There exists a growing demand for DNNs to simultaneously process multiple tasks~\cite{caruana97multitask, ruder2017overview}. In the speech domain, previous works attempt to train a single model to solve multiple tasks~\cite{chen2021speechnet,tjandra2017listening,ren2019almost,chen2015speech,kinnunen2017non, zhang2019joint} or use an auxiliary task to enhance the accuracy of a primary task~\cite{jain2018improved,chen2020aipnet,hsu2019disentangling,jia2018transfer, zhang2019non, zhang2019improving}. Motivated by the success of SSL in low-resource downstream tasks, multitask/multilingual learning has been adopted in both SSL pretraining~\cite{ravanelli2020multi, zaiem2021pretext, pascual2019learning, lu2022language, conneau2020unsupervised} to enrich learned speech representations, and finetuning scenarios~\cite{chen2021speech}. For example,
a recent relevant work~\cite{lu2022language} adopts language-adaptive pretraining on top of~\cite{conneau2020unsupervised}, where different sparse sub-networks are activated for different languages during multilingual pretraining and the gradients are accumulated on the shared super-network to learn better multilingual representations. In contrast, \METHOD{} is fundamentally different, as it targets the finetuning stage of a given pretrained speech SSL model and only tunes the connections of the model structure  \textit{without} tuning the SSL pretrained weights.

\textbf{ASR pruning.}
As ASR models become increasingly overparameterized for abstracting generalizable representations, pruning large-scale ASR models has drawn a growing attention. 
Early works trim down either the decoding search spaces~\cite{abdou2004beam,pylkkonen2005new,siivola2007growing,he2014reshaping,xu2018pruned,zhang2021tiny} or HMM state space~\cite{van1996adaptive}. Modern ASR paradigm has gradually shifted its focus to pruning end-to-end ASR models~\cite{yu2012exploiting, panchapagesan2021efficient, shangguan2019optimizing,wu2021dynamic, venkatesh2021memory,shi2021emformer,li2021efficient, braun2019parameter, narang2017exploring, gao2020rethinking, xue2013restructuring,povey2018semi}. Recently,~\cite{lai2021parp,prasad2022pada,zhao2021unified} prune speech SSL models towards more efficient low-resource ASR, e.g., PARP~\cite{lai2021parp} proposes a finetuning pipeline to identify the existence of lottery tickets within speech SSL models. In addition to the boosted pruning efficiency over PARP~\cite{lai2021parp}, \METHOD{} emphasizes the role of sparsity in encoding language-/task-specific information which can empower multitask/multilingual speech processing \textit{without} tuning the pretrained weights.

\begin{figure}[t!]
\centering
\includegraphics[width=0.95\textwidth]{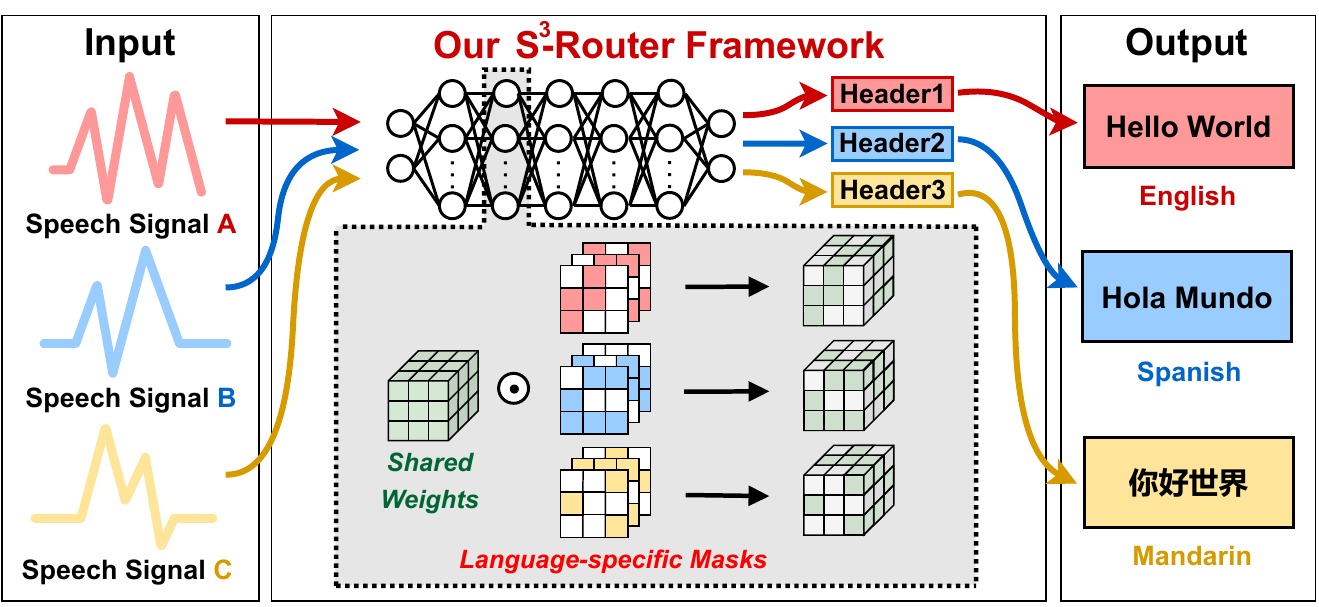}
\vspace{-0.7em}
\caption{An overview of our \METHOD{} framework, which receives multilingual speech signals denoted as A, B, and C here and then outputs the corresponding  text transcript of predication, based on one \textit{shared weight} model together with language-/task-specific \textit{binary} masks.}
\label{fig:overview}
\vspace{-1.5em}
\end{figure}

\section{The Proposed \METHOD{} Framework}
\label{sec:method}

\subsection{Drawn Inspirations from Previous Work}

Recent works~\cite{ramanujan2020s, malach2020proving,fu2021drawing} find that sub-networks featuring a decent inborn accuracy and adversarial robustness are hidden within randomly initialized networks \textit{without} any weight training. Specifically,~\cite{ramanujan2020s, malach2020proving} show that sub-networks with a decent accuracy, even matching that of their dense networks, can be identified from randomly initialized networks, and~\cite{fu2021drawing} shows an even stronger evidence: merely updating the sparsity patterns of model connections \textit{without} modifying the randomly initiated model weights can produce both accurate and adversarially robust models. 
These pioneering works imply that tuning the connections of the model structure, which can be characterized by the learned connection sparsity patterns, can be as effective as training the model weights. We hypothesize that \textbf{model sparsity not only can favor model efficiency, but also can serve as a similar role, i.e., another optimization knob, as model weights to encode language-/task-specific information}.

\subsection{Formulation and Optimization of \METHOD{}}
Inspired by the aforementioned intriguing hypothesis, we propose the \METHOD{} framework to empower efficient multilingual and multitask speech processing on top of SSL speech representations.                           

\textbf{Overview.}
As shown in Fig.~\ref{fig:overview}, given the raw audios of different spoken languages (or different tasks), \METHOD{} finetunes the connection patterns of the model structure for \textit{each} target spoken language/task via optimizing  language-/task-specific \textit{binary masks} on top of the \textit{shared weights} of a given speech SSL model, instead of finetuning the model weights as adopted in the common pretrain-finetune paradigm. Specifically, the learned binary masks of each language/task are multiplied with the shared model weights to mask out some connections within the given speech SSL model, where the remaining connections (or the induced connection sparsity) encode language-/task-specific information for different spoken languages and downstream tasks. Note that for each language or task, only one set of binary masks and one lightweight header, e.g., the classification head for ASR which is naturally non-shareable across languages due to diverse dictionary sizes, need to be independently trained, incurring negligible overhead (e.g.,$\leq$6.3\% storage of the whole backbone model).

\textbf{Formulation.}
Formally, our \METHOD{} framework can be formulated as:

\useshortskip
\begin{equation} \label{eq:objective}
    \argmin_{m_t} \,\, \sum_{(x_t,y_t)\in D_t}  \ell_t(f(m_t \odot \theta_{SSL}, x_t), y_t) \,\,\,\,\, s.t. \,\,\, ||m_t||_0  \leqslant k_t
\end{equation} 
\vspace{-1em}

\noindent where $(x_t,y_t)$ are the input audio and corresponding transcriptions/labels of a spoken language/task $t$ in a downstream dataset $D_t$, and $\theta_{SSL}$ is the SSL pretrained weights of the given speech SSL model $f$. Specifically, the mask set $m_t$ is applied on top of the model weights $\theta_{SSL}$, and optimized to minimize the loss function $l_t$, e.g., a CTC loss~\cite{graves2006connectionist} for ASR, subject to an $L_0$ sparsity constraint, where the number of non-zero elements in $m_t$ is limited to $k_t$, which serves as a hyperparameter for controlling and balancing (1) the amount of language-/task-specific information encoded in the connection sparsity and (2) model efficiency. Intuitively, if the sparsity is 0\% or 100\%, no new information is introduced during finetuning on the corresponding new/downstream language/task.

\textbf{Optimization.} To differentiably optimize $m_t$ in Eq.~(\ref{eq:objective}), we binarize $m_t$ and activate only its top $k_t$ elements during forward, while all the elements in $m_t$ are updated via straight-through estimation~\cite{bengio2013estimating} during backward. More specifically, during forward, we binarize $m_t$ to $\hat{m}_t$ via enforcing its top $k_t$ elements to 1 and other elements to 0, thus the forward function becomes $f(\hat{m}_t \odot \theta_{SSL}, x_t)$, which is the same for the inference process; during backward, we directly propagate the gradients from the binary mask $\hat m$ to $m$, i.e., $\frac{\partial l}{\partial m_t}\approx\frac{\partial l}{\partial \hat m_t}$, thus $m_t$ can be learned in a gradient-based manner.

\subsection{How to Initialize the Masks in \METHOD{}?}
\label{sec:mask_init}

\textbf{Importance of mask initialization.} A low learning rate is commonly adopted during finetuning to ensure effective inheritance of the SSL speech representations, resulting in merely smaller changes in each set of the masks $m_t$ in Eq.~(\ref{eq:objective}) as compared to training from scratch. Therefore, the mask initialization strategy plays an important role for the quality of the finally optimized masks in \METHOD{}. Next, we discuss the pros and cons of two intuitive mask initialization strategies:

\color{black}{\ding{172}} \textbf{Random initialization (RI).} We empirically find that adopting commonly used random initialization~\cite{he2015delving} in \METHOD{} can achieve a decent accuracy under a low sparsity. However, since no prior knowledge of the speech SSL model is utilized, the accuracy can largely drop with a high sparsity under random mask initialization, limiting extending \METHOD{} to be a practical pruning technique.

\color{black}{\ding{173}} \textbf{Weight magnitude based initialization (WMI).} As  weights magnitude can quantify the importance of weights as commonly used in pruning~\cite{han2015deep,han2015learning,li2016pruning,liu2018rethinking,frankle2018lottery}, taking the magnitudes of the given SSL pretrained weights $||\theta_{SSL}||$ in Eq.~\ref{eq:objective} as the initial values of masks might help utilize the knowledge learned during SSL pretraining. However, we empirically find that doing so causes worse trainability, i.e., the ranking of mask values is then more stable during finetuning than that under random initialization, and learned binary masks seldom change and mostly stick to their initial values. This may inhibit the optimization process and lead to sub-optimal learned masks.

\color{black}{\ding{174}} \textbf{Proposed Order-Preserving Random Initialization (ORI).} To marry the best of both above initialization strategies, we propose a new one, which can be viewed as a  weight rank based initialization. In ORI, we first acquire mask values of the same dimension as the shared weights via random initialization like~\cite{he2015delving}, and then perform magnitude-based sorting to assign a larger mask value to weight elements with a larger magnitude. Thus, the ranking order between the mask values of the weight elements is the same as that of their magnitudes. In this way, the mask trainability is maintained while the learned speech SSL model knowledge can be exploited (see Sec \ref{sec:exp_discard}).

\vspace{-0.5em}
\subsection{\METHOD{} is Useful in Various Application Scenarios}
\label{sec:applications}
\vspace{-0.5em}

\textbf{A new finetuning scheme.} \METHOD{} offers a new and equally effective finetuning scheme as it finetunes model connections given a speech SSL model.
As large-scale speech SSL models tend to overfit when finetuning their weights under a low-resource setting, we hypothesize that the binary optimization of the mask patterns in Eq.~(\ref{eq:objective}) can serve as regularization and thus alleviate overfitting.

\textbf{An efficient multilingual and multitask method.} \METHOD{} can naturally enable multilingual and multitask speech processing by merely switching among its learned language-/task-specific binary masks.
One advantage is that since the gradients of different spoken languages or tasks can now be independently accumulated on their corresponding masks \textit{without} interfering each other, the commonly observed gradient conflict issue~\cite{yu2020gradient} can be alleviated.

\textbf{A new pruning technique.} Since \METHOD{} encodes language/task-specific information via binary masks on top of shared model weights, it naturally introduces sparsity into the given model, and thus can naturally serve as a pruning technique. 
To further improve the achievable accuracy-efficiency trade-off and more fairly benchmark with SOTA ASR pruning methods, we propose a variant of \METHOD{} dubbed \METHOD{}-P, which first finetunes the model weights on the downstream audios and then prunes the model connections based on the learned binary masks in Eq.~(\ref{eq:objective}).

\textbf{An effective and simple tool to analyze what is encoded across language/task-specific speech SSL models.} Another exciting advantage of \METHOD{} is that it can provide a quantitative metric about how the pretained SSL speech presentation is utilized by different spoken languages/tasks via masking out languages/tasks-specific weights.

\section{\METHOD{}: Discarding $\leq$10\% Weights is All You Need}
\label{sec:exp_discard}

\subsection{Experiment Setup}
\label{sec:exp_setup} 
\vspace{-0.5em}
Here we evaluate \METHOD{} as a new finetuning scheme for downstream speech processing.

\textbf{Models.} We adopt wav2vec 2.0 base/large (wav2vec2-base/large)~\cite{baevski2020wav2vec} and data2vec~\cite{baevski2022data2vec} pretrained on LibriSpeech 960 hours~\cite{panayotov2015LibriSpeech} and xlsr~\cite{conneau2020unsupervised} pretrained on 128 languages sampled from CommonVoice~\cite{ardila2019common} as our speech SSL models.

\textbf{Datasets.} We consider 10 speech processing tasks, including low-resource English ASR on LibriSpeech~\cite{conneau2020unsupervised} with only 10min/1h/10h labeled data, following the dataset split in~\cite{baevski2020wav2vec,lai2021parp}, and low-resource phoneme recognition on CommonVoice~\cite{ardila2019common} with 1h labeled data per language, following the dataset split in~\cite{riviere2020unsupervised,conneau2020unsupervised}, as well as 8 speech processing tasks from SUPERB~\cite{yang2021superb}.

\textcolor{black}{\textbf{Finetuning settings.} Our code is built on top of fairseq~\cite{ott2019fairseq} and we follow the standard finetuning settings for each task, i.e., the default configurations in fairseq~\cite{ott2019fairseq} for ASR/phoneme recognition and those in SUPERB~\cite{yang2021superb} for other tasks. In particular, all our experiments on ASR/phoneme recognition adopt an Adam optimizer with an initial learning rate of 5e-5 plus a tri-stage schedule~\cite{baevski2020wav2vec} and we finetune wav2vec2-base/large for 12k/15k/20k steps on the 10m/1h/10h splits, respectively, and xlsr is finetuned for 12k steps for each spoken language. It takes about 10/24/24 GPU hours to finetune wav2vec2-base/large/xlsr for 12k steps with our \METHOD{}. We do not freeze all the layers except the final linear layer for the first 10k steps~\cite{baevski2020wav2vec}, following~\cite{lai2021parp}.} 

\textbf{\METHOD{} settings.} If not specifically stated, \METHOD{} only tunes the connections of (i.e., applies learnable masks on) the feed-forward networks (FFNs) of transformer structures~\cite{vaswani2017attention} and fixes all other weights, which is empirically found to be optimal based on the ablation studies in Sec.~\ref{sec:exp_ablation}. In addition, if not stated, we adopt our ORI mask initialization by default and do not apply language models for a fair benchmark of ASR performance, following~\cite{lai2021parp}.

\begin{figure}[t!]
  \vspace{-0.3em}
\centering
\includegraphics[width=\linewidth]{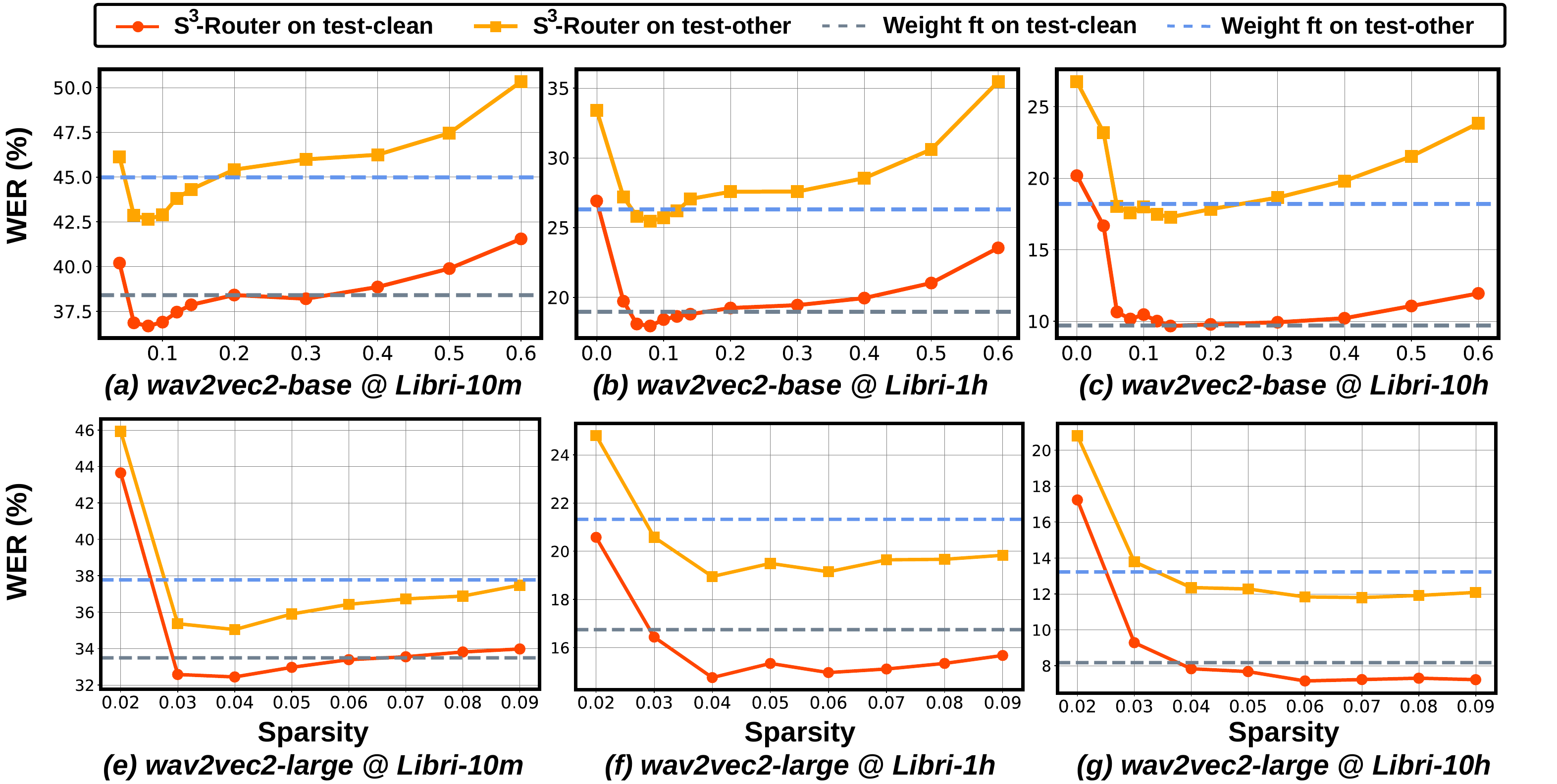}
\vspace{-1.5em}
\caption{Benchmark our \METHOD{} and standard weight finetuning on the test-clean/test-other sets of LibriSpeech on top of wav2vec2-base/large under different low-resource settings. }
\label{fig:ft_wav2vec}
\end{figure}

\begin{figure}[t!]
  \vspace{-1em}
\centering
\includegraphics[width=\linewidth]{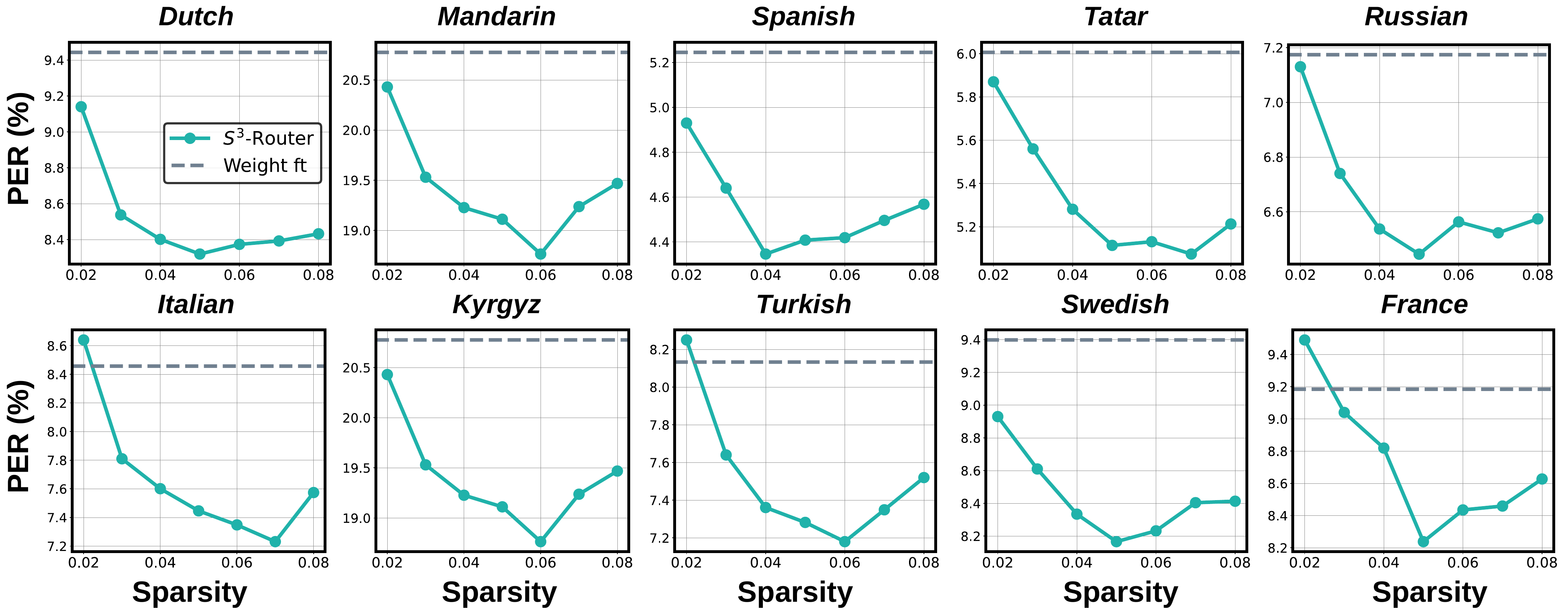}
\vspace{-1.5em}
\caption{Benchmark our \METHOD{} and weight finetuning on xlsr across 10 spoken languages.}
\label{fig:ft_xlsr_cv}
\vspace{-1.5em}
\end{figure}

\subsection{Benchmark on Low-resource English ASR}
\label{sec:exp_english_asr}
\textbf{Finetuning on wav2vec2-base.} We apply our \METHOD{} on wav2vec2-base under different low-resources settings as shown in Fig.~\ref{fig:ft_wav2vec} (a)$\sim$(c). We can observe that our \METHOD{} can consistently outperform the standard weight finetuning in terms of the achievable WER, i.e., the lowest WER at the corresponding optimal sparsity. In particular, our \METHOD{} achieves a 1.72\%/2.34\% reduction in word error rate (WER) under a sparsity ratio of 8\% on the test-clean/test-other set of LibriSpeech, respectively, when being finetuned on 10min labeled data. This indicates that finetuning connections via our \METHOD{} can be a competitive alternative with reduced WER for finetuning weights under low-resource settings, which provides a new paradigm for finetuning speech SSL models.

\textbf{Ablation studies of mask initialization.} We equip our \METHOD{} with different mask initialization schemes in Sec.~\ref{sec:mask_init} and show their achievable WER on low-resource English ASR in Tab.~\ref{tab:mask_init}. We can observe that \underline{(1)} the proposed ORI mask initialization consistently outperforms the other two schemes in terms of the achievable WER, and \underline{(2)} even random mask initialization can match or surpass the performance of standard weight finetuning.
We also provide their complete sparsity-WER trade-offs in the appendix and find that  weight magnitude based initialization favors larger sparsity ratios, while it is harder to overturn the ranking between masks via gradients, resulting in inferior achievable WER.

\begin{wraptable}{r}{0.5\textwidth}
  \vspace{-1em}
  \centering
  \caption{Benchmark different mask initialization schemes of \METHOD{} with weight finetuning on LibriSpeech test-clean/test-other sets.}
    \resizebox{0.5\textwidth}{!}{
\begin{tabular}{cccc}
\toprule
\textbf{Method}     & \textbf{Libri-10m}   & \textbf{Libri-1h}    & \textbf{Libri-10h}  \\ \midrule
Weight ft           & 38.40/44.98          & 18.963/26.298        & 9.7/18.19           \\ \midrule
RI                  & 36.98/43.42          & 18.19/26.31          & 9.78/18.17          \\
WMI                 & 40.09/46.63          & 19.99/27.16          & 11.17/18.97         \\
\textbf{ORI (Ours)} & \textbf{36.68/42.64} & \textbf{17.95/25.47} & \textbf{9.66/17.27} \\ \bottomrule
\end{tabular}
    }
  \vspace{-1.5em}
  \label{tab:mask_init}%
\end{wraptable}%

\textbf{Scalability to larger speech SSL models.} We apply \METHOD{} to a larger model wav2vec-large. As shown in Fig.~\ref{fig:ft_wav2vec} (e)$\sim$(g), we can see that the achievable WER of \METHOD{} still consistently outperforms the standard weight finetuning across all downstream datasets, e.g.,  a 1.99\%/2.37\% WER reduction under a 4\% sparsity on the test-clean/test-other set, respectively, when being finetuned on the 1h labeled data. Consistent results on xlsr are provided in the appendix.

\underline{Insights.} 
This set of experiments indicates that \color{black}{\ding{182}} our \METHOD{} features a good scalability to larger speech SSL models as well as a good generality for different pretraining schemes, and \color{black}{\ding{183}} our method effectively reduces the overfitting on more overparameterized speech SSL models according to the larger performance gains, thanks to the regularization effect of the binary optimization process in Eq.~(\ref{eq:objective}), making our method an appealing solution for more advanced speech SSL models.

\begin{wraptable}{r}{0.5\textwidth}
  \vspace{-2em}
  \centering
  \caption{Benchmark our \METHOD{} with standard weight finetuning when being equipped with a 4-gram LM~\cite{heafield2013scalable}.}
    \resizebox{0.5\textwidth}{!}{
\begin{tabular}{ccccc}
\toprule
\textbf{Model} & \textbf{Method}   & \textbf{Libri-10m}   & \textbf{Libri-1h}    & \textbf{Libri-10h}  \\ \midrule
\multirow{2}{*}{\begin{tabular}[c]{@{}c@{}}wav2vec2\\-base\end{tabular}}  & Weight ft & 26.55/32.95 & 11.54/18.45 & 6.65/13.97 \\
               & \textbf{\METHOD{}} & \textbf{25.28/31.07} & \textbf{11.28/18.33} & \textbf{6.42/13.14} \\  \midrule
\multirow{2}{*}{\begin{tabular}[c]{@{}c@{}}wav2vec2\\-large\end{tabular}} & Weight ft & 24.83/29.05 & 11.18/15.80 & 5.52/10.31 \\
               & \textbf{\METHOD{}} & \textbf{23.17/25.83} & \textbf{9.74/13.77}  & \textbf{4.98/9.27}  \\ \bottomrule
\end{tabular}
    }
  \vspace{-1.5em}
  \label{tab:lm}%
\end{wraptable}%

\textbf{Add language models (LMs).}  We further apply a 4-gram LM~\cite{heafield2013scalable} as the decoder for both our \METHOD{} at the optimal sparsity ratio in Fig.~\ref{fig:ft_wav2vec} and the weight finetuning baselines. 
In particular, we adopt the official 4-gram language model (LM)~\cite{heafield2013scalable} with a beam size of 50, an LM weight of 2, and a word insertion penalty of -1.
As shown in Tab.~\ref{tab:lm}, our \METHOD{} still consistently outperforms weight finetuning with more notable reductions on wav2vec2-large, e.g., a 1.66\%/3.22\% WER reduction on the test-clean/test-other set of LibriSpeech when being finetuned on 10min labeled data.

\begin{wraptable}{r}{0.5\textwidth}
  \vspace{-2em}
  \centering
  \caption{Benchmark our \METHOD{} under different sparsity ratios with weight finetuning on top of data2vec on LibriSpeech test-clean/test-other sets.}
    \resizebox{0.5\textwidth}{!}{
\begin{tabular}{cccc}
\toprule
\textbf{Method} & \textbf{Libri-10m} & \textbf{Libri-1h} & \textbf{Libri-10h} \\ \midrule
Standard ft & 30.75/34.612 & 14.15/19.61 & 7.28/13.11 \\ \midrule \midrule
\METHOD{}@0.07 & 30.78/35.17 & 14.09/19.72 & 7.56/13.39 \\
\METHOD{}@0.08 & 30.70/34.45 & 13.96/19.41 & 7.43/13.34 \\
\METHOD{}@0.09 & \textbf{29.86/34.09} & \textbf{13.92/19.43} & 7.23/13.25 \\
\METHOD{}@0.10 & 31.30/35.07 & 14.27/20.10 & \textbf{7.05/12.98}  \\ \bottomrule
\end{tabular}
    }
  \vspace{-1em}
  \label{tab:data2vec}%
\end{wraptable}%

\textbf{Scalability to speech models pretrained by other SSL paradigms.} 
To validate the generalization capability of our \METHOD{} across speech models pretrained by other SSL paradigms, we further apply our method on the SOTA speech SSL model data2vec~\cite{baevski2022data2vec} under different sparsity ratios, which features a new SSL pretraining paradigm based on self-distillation. Here we follow the same finetuning setting as wav2vec2-base, which is also the default finetuning setting of data2vec in fairseq~\cite{ott2019fairseq}.
As shown in Tab.~\ref{tab:data2vec}, we can observe that our \METHOD{} still achieves lower WER across all resource settings, e.g., a 0.89\% WER reduction on LibriSpeech test-clean when being trained with 10m labeled speech. This indicates that our \METHOD{} can generally serve as a competitive finetuning paradigm independent of the SSL scheme.

\vspace{-0.5em}
\subsection{Benchmark on Low-resource Cross-lingual Transfer}
\label{sec:exp_cross_lingual}
\vspace{-0.5em}

We further evaluate our \METHOD{} under two cross-lingual transfer settings, including (1) finetuning the multilingual pretrained xlsr on different low-resource languages in CommonVoice~\cite{ardila2019common}, and (2) a high-to-low resource transfer setting where the monolingual (English) pretrained wav2vec2-base is finetuned on multiple low-resource languages in CommonVoice.

\textbf{Cross-lingual transfer on xlsr.} As shown in Fig.~\ref{fig:ft_xlsr_cv}, our \METHOD{} consistently outperforms standard weight finetuning in terms of the achievable phoneme error rate (PER) across all the 10 spoken languages, e.g., a 1.12\%/2.01\% PER reduction on Dutch/Mandarin, respectively.

\begin{wraptable}{r}{0.5\textwidth}
  \vspace{-1.5em}
  \centering
  \caption{Benchmark our \METHOD{} and weight finetuning on wav2vec2-base and CommonVoice.}
    \resizebox{0.5\textwidth}{!}{
\begin{tabular}{cccccc}
\toprule
\textbf{Language}          & \textbf{Dutch}   & \textbf{Mandarin} & \textbf{Spanish} & \textbf{Tatar}   & \textbf{Russian} \\ \midrule
Weight ft                  & 19.82           & 26.67            & 13.86          & 11.14           & 17.05           \\
\METHOD{} & \textbf{18.51}  & \textbf{26.10}   & \textbf{13.37}  & \textbf{10.94}  & \textbf{16.33}  \\ \midrule
\textbf{Language}          & \textbf{Italian} & \textbf{Kyrgyz}   & \textbf{Turkish} & \textbf{Swedish} & \textbf{France}  \\ \midrule
Weight ft                  & 19.27           & 13.41            & 15.70           & 20.81           & 19.35           \\
\METHOD{} & \textbf{18.29}  & \textbf{12.30}   & \textbf{14.82}  & \textbf{19.64}  & \textbf{17.94}  \\ \bottomrule
\end{tabular}
    }
  \vspace{-1.5em}
  \label{tab:ft_wav2vec2_cv}%
\end{wraptable}%

\textbf{High-to-low resource transfer on wav2vec2-base.} As shown in Tab.~\ref{tab:ft_wav2vec2_cv}, on top of the English pretrained wav2vec2-base, finetuning the connections with \METHOD{} still wins the lowest achievable PER over the baseline.
Consistent results on wav2vec2-large are in the appendix.

\underline{Insights.} This set of experiments implies that for each downstream spoken language, there exist decent subnetworks for processing its language-specific information even in monolingual pretrained speech SSL models. In another words, properly learned sparsity can encode language-specific information with competitive ASR performance.

\subsection{Benchmark on More Downstream Speech Processing Tasks}
\label{sec:exp_superb}

We further evaluate our \METHOD{} via finetuning wav2vec2-base on 8 speech processing tasks from SUPERB~\cite{yang2021superb}, covering different aspects of speech (content/speaker/semantics/paralinguistics). 
We show the achievable task performance as well as the corresponding optimal sparsity ratios in Tab.~\ref{tab:superb} and find that our method surpasses the task performance over standard weight finetuning across all 8 tasks via discarding $\leq$10\% weights, indicating that properly learned sparsity can also effectively encode task-specific information. We provide the sparsity-performance trade-offs in the appendix.

\begin{table}[t]
  \centering
  \caption{Benchmark \METHOD{} with standard weight finetuning on 8 tasks from SUPERB~\cite{yang2021superb}.}
  \vspace{-0.5em}
    \resizebox{0.98\textwidth}{!}{
\begin{tabular}{c|c|ccc|c|ccc}
\toprule
\textbf{Catagory}          & \textbf{Content} & \multicolumn{3}{c|}{\textbf{Speaker}}          & \textbf{Paralinguistics} & \multicolumn{3}{c}{\textbf{Semantics}}          \\ \midrule
\textbf{Task} &
  \textbf{\begin{tabular}[c]{@{}c@{}}Keyword \\ Spotting\end{tabular}} &
  \textbf{\begin{tabular}[c]{@{}c@{}}Speaker \\ Identification\end{tabular}} &
  \textbf{\begin{tabular}[c]{@{}c@{}}Speaker \\ Verification\end{tabular}} &
  \textbf{\begin{tabular}[c]{@{}c@{}}Speaker \\ Diarization\end{tabular}} &
  \textbf{\begin{tabular}[c]{@{}c@{}}Emotion \\ Recognition\end{tabular}} &
  \textbf{\begin{tabular}[c]{@{}c@{}}Intent \\ Classification\end{tabular}} &
  \textbf{\begin{tabular}[c]{@{}c@{}}Slot \\ Filling\end{tabular}} &
  \textbf{\begin{tabular}[c]{@{}c@{}}Speech \\ Translation\end{tabular}} \\ \midrule
Metric            & Acc $\uparrow$     & Acc $\uparrow$   & EER $\downarrow$ & DER $\downarrow$  & Acc $\uparrow$         & Acc $\uparrow$  & F1 $\uparrow$   & BLEU $\uparrow$  \\ \midrule
Weight ft                  & 95.5             & 66.8           &   7.24           & 7.41          & 61.5                     & 91.48         & 88.1           & 18.85          \\
\textbf{\METHOD{}} & \textbf{95.7}    & \textbf{71.07} & \textbf{6.79}    & \textbf{7.18} & \textbf{62.13}           & \textbf{92.6} & \textbf{88.76} & \textbf{19.01} \\ \midrule
Opt Spar.                  & 0.1              & 0.06           & 0.1          & 0.08          & 0.1                      & 0.1           & 0.1            & 0.08        \\ \bottomrule  
\end{tabular}
    }
  \vspace{-1em}
  \label{tab:superb}%
\end{table}%

\subsection{Empowering Multilingual and Multitask Speech Processing}
\label{sec:exp_multilingual}
Since all the aforementioned results achieved on the same speech SSL model via  \METHOD{} share pretrained weights, it can simultaneously enable multilingual and multitask speech processing, thanks to the independent accumulation of task-specific gradients that avoids gradient conflicts~\cite{yu2020gradient}, of superior scalability with the number of tasks. For example, as compared to independent weight finetuning for each language/task, \METHOD{} can simultaneously support 11 languages in Sec.~\ref{sec:exp_english_asr}/~\ref{sec:exp_cross_lingual} and 8 tasks in Sec.~\ref{sec:exp_superb} using one wav2vec2-base while achieving a win-win in both accuracy (as validated in Sec.~\ref{sec:exp_english_asr}/~\ref{sec:exp_cross_lingual}/~\ref{sec:exp_superb}) and efficiency, i.e., more than 88.5\% reductions in model parameters.

\subsection{Benchmark \METHOD{} with Adaptor Tuning} 

\begin{wraptable}{r}{0.5\textwidth}
  \vspace{-2em}
  \centering
  \caption{Benchmark our \METHOD{} with adaptor tuning~\cite{le2021lightweight} on top of wav2vec2-base across LibriSpeech test-clean/other and CommonVoice.}
    \resizebox{0.5\textwidth}{!}{
\begin{tabular}{cccc}
\toprule
\textbf{Method} & \textbf{Libri-10m} & \textbf{Libri-1h} & \textbf{Libri-10h} \\ \midrule
Standard ft & 38.40/44.98 & 18.96/26.30 & 9.70/18.19 \\
Adaptor~\cite{le2021lightweight} & 41.82/49.01 & 21.19/28.92 & 12.26/19.67 \\
\METHOD{} & \textbf{36.68/42.64} & \textbf{17.95/25.47} & \textbf{9.77/17.27} \\ \midrule \midrule
\textbf{Method} & \textbf{Dutch} & \textbf{Spanish} & \textbf{Mandarin} \\ \midrule
Standard ft & 19.82 & 13.86 & 26.67 \\
Adaptor~\cite{le2021lightweight} & 22.63 & 15.89 & 29.03 \\
\METHOD{} & \textbf{18.51} & \textbf{13.37} & \textbf{26.10} \\ \bottomrule
\end{tabular}
    }
  \vspace{-1em}
  \label{tab:adaptor}%
\end{wraptable}%

We further benchmark our \METHOD{} with adaptor tuning~\cite{zhao2020masking}, which has emerged as an alternative finetuning scheme. In particular, we reproduce the speech adaptor design in~\cite{le2021lightweight}, following their open-sourced implementation, on wav2vec2-base and benchmark with our reported results of \METHOD{} under the best sparsity settings for different languages. As shown in Tab.~\ref{tab:adaptor}, we can observe that our \METHOD{} still consistently achieves the lowest WER/PER across all languages and resource settings, e.g., a 5.14\%/3.24\% WER reduction on LibriSpeech test-clean when being trained on LibriSpeech-10m/1h, respectively.

\underline{Insight.} Under a low-resource setting, it is hard for adaptor tuning to maintain a comparable accuracy over standard weight finetuning, while our \METHOD{} often outperforms standard weight finetuning as shown in Sec.~\ref{sec:exp_english_asr}, thanks to its binary optimization process on the masks, which can potentially regularize the learning process and lead to better downstream performances as compared to finetuning the overparameterized model weights. This indicates that \METHOD{} is a better finetuning scheme over adaptor tuning, especially under low-resource settings.

\begin{figure}[h!]
  \vspace{-1em}
\centering
\includegraphics[width=\linewidth]{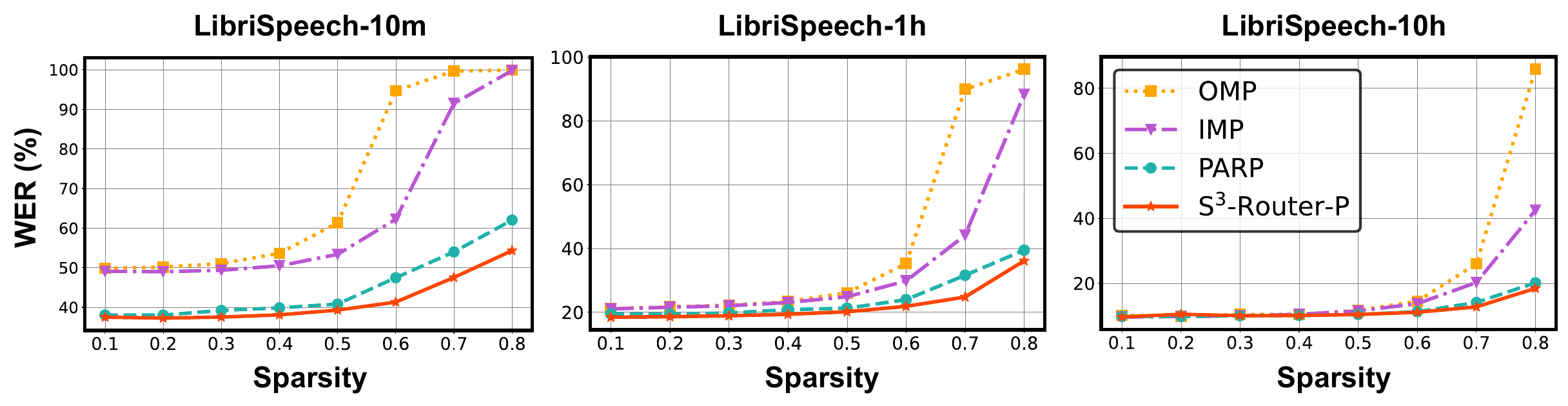}
\vspace{-2em}
\caption{Benchmark our \METHOD{}-P against OMP, IMP, and PARP~\cite{lai2021parp} for pruning wav2vec2-base on LibriSpeech. The WER on the test-clean set is reported.}
\label{fig:prune_libri}
\vspace{-1.5em}
\end{figure}

\section{\METHOD{}-P: Pruning ASR Models for Enhancing Efficiency}
\textbf{Setup.}
For evaluating \METHOD{}-P (see Sec.~\ref{sec:applications}), we conduct two modifications: \underline{(1)} both FFN and self-attention (SA) modules are pruned for a fair comparison, and \underline{(2)} we adopt weight magnitude based mask initialization for two reasons: firstly, it features the best scalability to high sparsity among the three initialization schemes; secondly, since the weights have been finetuned, their magnitudes can serve to indicate their importance on the downstream speech. Ablation studies for pruning under different mask initialization schemes are in the appendix.

\begin{wrapfigure}{r}{0.35\textwidth} 
    \centering
    \vspace{-1.5em}
    \includegraphics[width=0.3\textwidth]{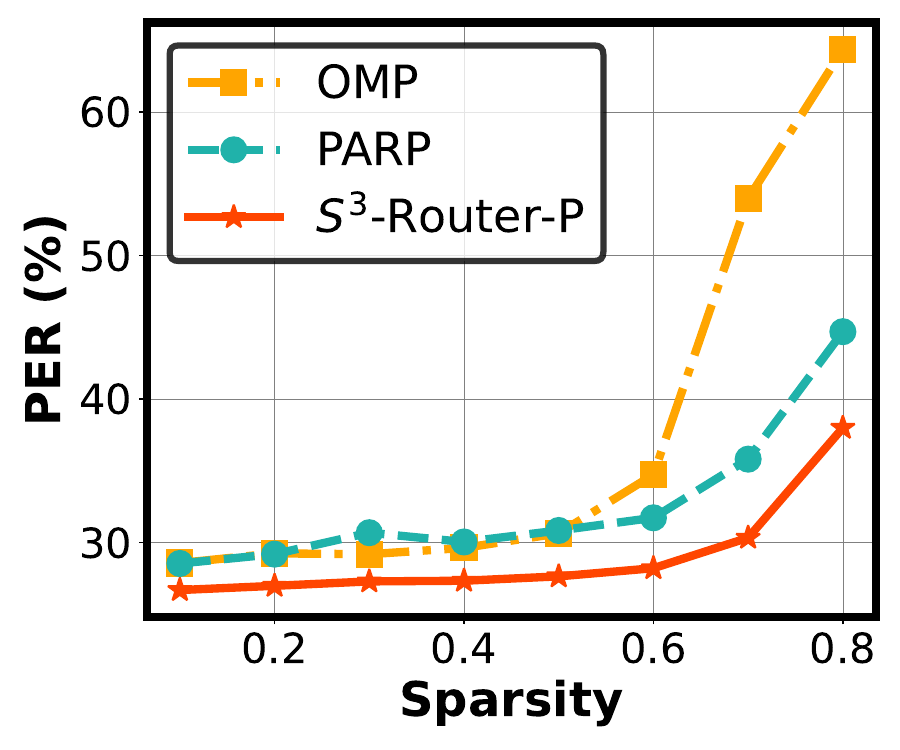}
    \vspace{-1em}
    \caption{Benchmark our \METHOD{}-P against OMP and PARP for pruning on Mandarin.}
    \label{fig:prune_cv}
    \vspace{-1.5em}
\end{wrapfigure}
For our baselines, we benchmark with three ASR pruning methods, i.e., one-shot/iterative magnitude pruning (OMP/IMP), and the SOTA method PARP~\cite{lai2021parp} under the best setting (dubbed PARP-P), and directly adopt their reported results in~\cite{lai2021parp}.

\textbf{Pruning wav2vec2-base on LibriSpeech.} As shown in Fig.~\ref{fig:prune_libri}, we can observe that \underline{(1)} our \METHOD{}-P consistently achieves the most competitive WER-sparsity trade-offs across the two models and three datasets, e.g., a 6.46\% lower WER over PARP under a sparsity ratio of 0.7 with 10min labeled data, and \underline{(2)} our method features a better scalability to more stringent low-resource scenarios according to the large performance gains on LibriSpeech-10m.

\textbf{Pruning wav2vec2-base on Mandarin@CommonVoice.} As shown in Fig.~\ref{fig:prune_cv}, consistent observations can be drawn that our method still wins the WER-sparsity trade-offs.
More pruning results are provided in the appendix.

\vspace{-0.5em}
\section{\METHOD{}: What is Encoded in Speech SSL Models?}
\label{sec:exp_ablation}

\subsection{What is The Roles of Different Modules in Speech SSL Models?}
We tune the connections of a subset of the modules in the speech SSL model via our \METHOD{} to study their contributions with other modules fixed in terms of both weights and connections.

\begin{wraptable}{r}{0.5\textwidth}
  \vspace{-1.8em}
  \centering
  \caption{The achievable WER on LibriSpeech test-clean/test-other when finetuning different modules.}
    \resizebox{0.5\textwidth}{!}{
\begin{tabular}{cccc}
\toprule
Sparsity & SA only       & FFN only      & Both          \\ \midrule
0.1      & 24.39/30.79 & \textbf{18.62/26.12} & 19.26/27.36 \\
0.2      & 23.06/30.22 & \textbf{19.54/27.07}  & 18.99/27.38 \\
0.3      & 22.78/30.21  & \textbf{19.59/27.58} & 19.65/28.40 \\
0.4      & 23.22/31.28   & 20.02/28.97  & \textbf{19.78/27.98} \\ \bottomrule
\end{tabular}
    }
  \vspace{-1em}
  \label{tab:ffn_sa}%
\end{wraptable}%

\textbf{FFN vs. SA.} We tune the connections of FFN, SA, or both on top of wav2vec2-base and LibriSpeech-1h as shown in Tab.~\ref{tab:ffn_sa}. We can observe that tuning the connections of FFN only wins the lowest achievable WER, even outperforming tuning both FFN and SA. This indicates that the SSL pretrained SA modules are generalizable enough to capture temporal relationship between tokens for downstream speech thus only FFN needs to be tuned for encoding new information about downstream tasks.

\begin{wraptable}{r}{0.4\textwidth}
  \vspace{-2em}
  \centering
  \caption{Apply \METHOD{} on different groups of blocks. WERs on LibriSpeech test-clean/test-other are reported.}
    \resizebox{0.4\textwidth}{!}{
\begin{tabular}{ccc}
\toprule
\textbf{Group} & \textbf{Libri-1h} & \textbf{Libri-10h} \\ \midrule
{[}1,0,0,0{]} & 85.68/92.12 & 77.65/86.12 \\
{[}0,1,0,0{]} & 54.96/67.17 & 42.26/56.24 \\
{[}0,0,1,0{]} & \textbf{24.57/32.68} & \textbf{14.95/22.96} \\
{[}0,0,0,1{]} & 35.12/42.39 & 20.88/27.29 \\ \bottomrule
\end{tabular}
    }
  \vspace{-1.5em}
  \label{tab:role_block}%
\end{wraptable}%

\textbf{Roles of different blocks.} To study the roles of the blocks at different depths, we uniformly divide the 12 blocks in wav2vec2-base into 4 groups and only tune the connections of one group. As shown in Tab.~\ref{tab:role_block}, we indicate the tunable group as 1 and other fixed ones as 0. We can observe that generally tuning later groups achieves lower WER and tuning the 1st/2nd groups only will result in unacceptably high WER, aligning with our intuition that early blocks in speech SSL models extract generalizable phonetic features while later blocks capture task-specific features thus are required to be tuned. In addition, tuning the 3rd group achieves the lowest WER, even close to that of tuning all groups, which we assume is because the features starting from the 3rd group are task-specific and not generalizable across downstream tasks thus demandingly require to be finetuned.

\begin{wraptable}{r}{0.5\textwidth}
  \vspace{-0.8em}
  \centering
  \vspace{-1em}
  \caption{The optimal sparsity ratios for the lowest WER across different models and resources.}
  \vspace{0.3em}
    \resizebox{0.5\textwidth}{!}{
\begin{tabular}{cccc}
\toprule
\textbf{Model}          & \textbf{Libri-10m} & \textbf{Libri-1h} & \textbf{Libri-10h} \\ \midrule
wav2vec2-base  & 0.08      & 0.10     & 0.20      \\
wav2vec2-large & 0.04      & 0.04     & 0.06      \\
xlsr           & 0.04      & 0.05     & 0.07     \\ \bottomrule
\end{tabular}
    }
  \vspace{-1em}
  \label{tab:opt_sparsity}%
\end{wraptable}%

\vspace{-0.5em}
\subsection{How Much New Information is Learned by Finetuning?}
\vspace{-0.5em}
We visualize the optimal sparsity ratio for achieving the lowest WER on  LibriSpeech-10m/1h/10h, which serves as an indicator about the amount of new information learned by finetuning, across wav2vec2-base/large/xlsr in Tab.~\ref{tab:opt_sparsity}. We can observe that \underline{(1)} larger speech SSL models reach their optimal performance under lower sparsity, i.e., relatively less amount of tuning could lead to their decent downstream performance thanks to their overparameterization, and \underline{(2)} finetuning on more resources leads to higher optimal sparsity, indicating that speech SSL models have more confidence to update the SSL pretrained representations given more resources.

\subsection{How is The Learned Masks Correlated to Phonetics?}
\label{sec:exp_correlation}
    \vspace{-0.5em}

Given the decent performance achieved by the learned masks, we are interested their correlations with human expertise in phonetics. In particular, we wonder whether the similarity between the learned masks of two languages aligns with that between their phoneme inventories.

\textbf{Setup.} Given the 11 languages adopted in Sec.~\ref{sec:exp_english_asr} and~\ref{sec:exp_cross_lingual}, we calculate layer-wise cosine similarities of their learned masks, under a sparsity ratio of 0.1 with near-optimal performance, between each pair of them. We pick the mask similarity of the first layer as our metric without losing generality. 
We further measure the cosine similarity between their corresponding phoneme inventories  acquired from Phoible~\cite{phoible}, a cross-linguistic phonological inventory database which covers over 2000 languages.
Following~\cite{gao2021zero}, we combine the inventories for all languages
to a shared phoneme inventory and use a binary vector to
indicate the phone inventory of each language, thus the language-wise phonetic similarities can be calculated as the cosine similarities between their corresponding binary vectors.

\begin{wrapfigure}{r}{0.5\textwidth}
    \centering
    \vspace{-1.5em}
    \includegraphics[width=0.5\textwidth]{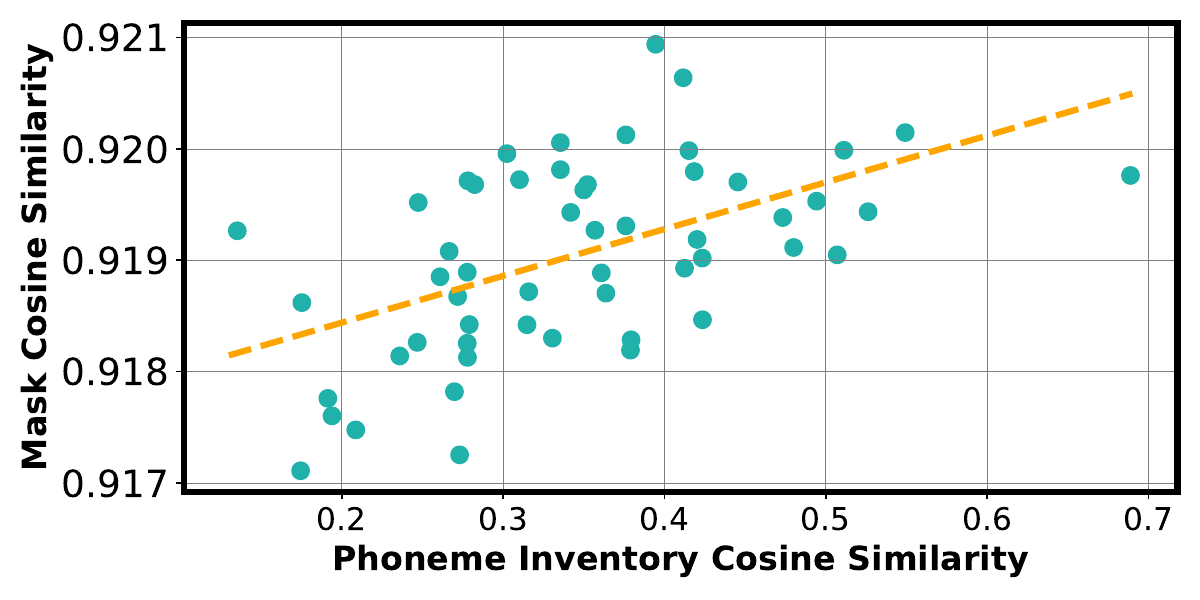}
    \vspace{-2em}
    \caption{Visualizing the correlation between the similarity of the learned masks and that of the corresponding phoneme inventories of language pairs.}
    \label{fig:correlation}
    \vspace{-1em}
\end{wrapfigure}

\textbf{Results and analysis.} We visualize the two similarity metrics of each language pair in Fig.~\ref{fig:correlation} and find that the Pearson Correlation Coefficient~\cite{benesty2009pearson} and the Spearman correlation coefficient~\cite{sedgwick2014spearman} between the two similarity metrics are 0.527 and 0.548, respectively. This indicates that the similarities of our learned masks are non-trivially correlated with that of phonetics while the former also provides new insights in the eyes of speech SSL models, which could shed light on future advances in zero-shot cross-lingual transfer~\cite{gao2021zero}. We provide more insightful correlation analysis as well as the visualization of layer-wise mask similarities between different languages in the appendix.

\vspace{-0.5em}
\section{Conclusion}
\label{sec:conclusion}
\vspace{-0.5em}
Motivating by the empirical success of SSL speech representations in low-resource speech processing, we propose \METHOD{} to facilitate the practical usage of speech SSL models via finetuning their connections instead of weights and encoding language-/task-specific information via sparsity. Extensive experiments validate that \METHOD{} not only serves as a stronger alternative with alleviated overfitting and enhanced accuracy for standard weight finetuning, but also empowers efficient multilingual and multitask speech processing. Our insights could enhance our understandings about what is encoded in speech SSL models and thus shed light on future advances in SSL speech representations.

\section*{Acknowledgements}
\vspace{-0.5em}
The work is supported by the National Science Foundation (NSF) through the CCRI program (Award number: 2016727) and an IBM faculty award received by Dr. Yingyan Lin.

\bibliographystyle{unsrt}
\bibliography{ref}

\section*{Checklist}

\begin{enumerate}

\item For all authors...
\begin{enumerate}
  \item Do the main claims made in the abstract and introduction accurately reflect the paper's contributions and scope?
    \answerYes{}
  \item Did you describe the limitations of your work?
    \answerYes{} In Sec.~\ref{sec:exp_correlation}, we discuss the potential of supporting zero-shot cross-lingual transfer as our future work, which is a new problem created by our work.
  \item Did you discuss any potential negative societal impacts of your work?
    \answerNo{} Our work proposes a framework to empower efficient multilingual and multitask speech processing, which does not notably correlate with any negative societal impacts.
  \item Have you read the ethics review guidelines and ensured that your paper conforms to them?
    \answerYes{}
\end{enumerate}

\item If you are including theoretical results...
\begin{enumerate}
  \item Did you state the full set of assumptions of all theoretical results?
    \answerNA{}
        \item Did you include complete proofs of all theoretical results?
    \answerNA{}
\end{enumerate}

\item If you ran experiments...
\begin{enumerate}
  \item Did you include the code, data, and instructions needed to reproduce the main experimental results (either in the supplemental material or as a URL)?
    \answerNo{} We will release our code, data, and instructions upon acceptance so that we can ensure ease of use.
  \item Did you specify all the training details (e.g., data splits, hyperparameters, how they were chosen)?
\answerYes{} See Sec.~\ref{sec:exp_setup} and the appendix.
        \item Did you report error bars (e.g., with respect to the random seed after running experiments multiple times)?
    \answerNo{}  We report the average results for clarity of the figures.
        \item Did you include the total amount of compute and the type of resources used (e.g., type of GPUs, internal cluster, or cloud provider)?
    \answerNo{} because the experiments are not computationally intensive.
\end{enumerate}

\item If you are using existing assets (e.g., code, data, models) or curating/releasing new assets...
\begin{enumerate}
  \item If your work uses existing assets, did you cite the creators?
    \answerYes{}
  \item Did you mention the license of the assets?
    \answerNo{} The used dataset and the license information can be found via the cited reference.
  \item Did you include any new assets either in the supplemental material or as a URL?
    \answerNo{}
  \item Did you discuss whether and how consent was obtained from people whose data you're using/curating?
    \answerNo{}
  \item Did you discuss whether the data you are using/curating contains personally identifiable information or offensive content?
    \answerNo{}
\end{enumerate}

\item If you used crowdsourcing or conducted research with human subjects...
\begin{enumerate}
  \item Did you include the full text of instructions given to participants and screenshots, if applicable?
    \answerNA{}
  \item Did you describe any potential participant risks, with links to Institutional Review Board (IRB) approvals, if applicable?
    \answerNA{}
  \item Did you include the estimated hourly wage paid to participants and the total amount spent on participant compensation?
    \answerNA{}
\end{enumerate}

\end{enumerate}


\end{document}


\maketitle

\section{Overview and Outline}
\label{sec:overview}
In this supplement, we provide more experiments and analysis as a complement to the main content, which are outlined below: 

\begin{itemize}
    \item We provide more results of our \METHOD{} as a new finetuning paradigm in Sec.~\ref{sec:finetuning}, including xlsr on LibriSpeech and wav2vec2-large on CommonVoice, and the sparsity-WER trade-offs achieved by different mask initialization schemes or on different speech processing tasks;
    \item We provide more results of our \METHOD{} for ASR pruning in Sec.~\ref{sec:pruning}, including an ablation study of different mask initialization schemes and more benchmarks with PARP~\cite{lai2021parp} for ASR pruning on more speech SSL models;
    \item We study more properties of speech SSL models via our \METHOD{} in Sec.~\ref{sec:correlation}, including the layer-wise similarity of learned masks between different languages and more analysis regarding the correlation between the learned masks of and phonetics;
\end{itemize}

\section{More Evaluation Results of \METHOD{} as a New Finetuning Paradigm}
\label{sec:finetuning}

\begin{figure}[t!]
  \vspace{-0.3em}
\centering
\includegraphics[width=\linewidth]{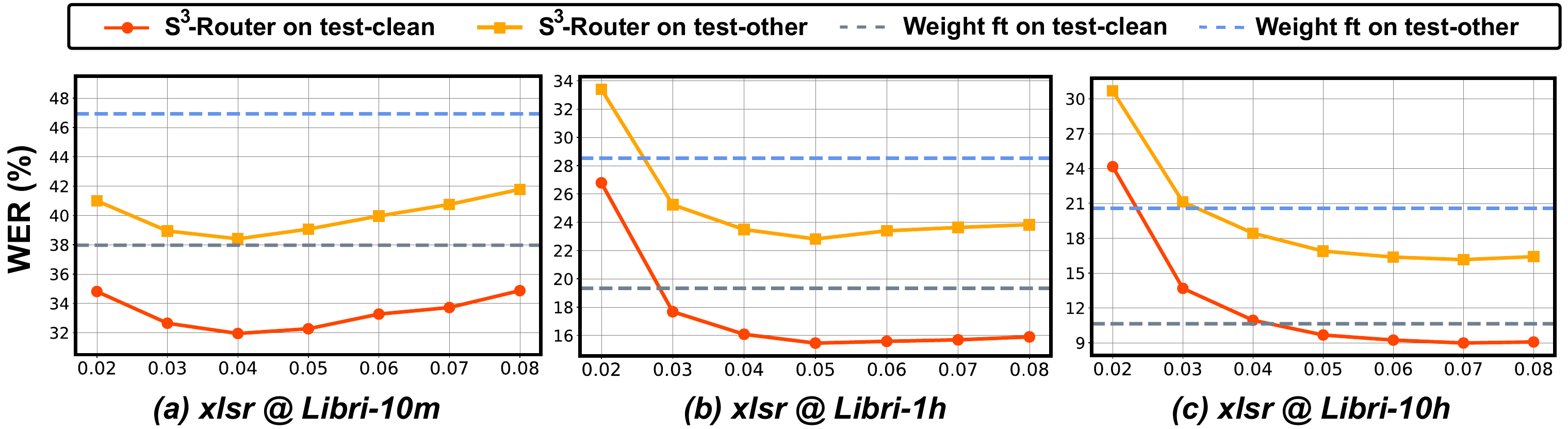}
\vspace{-1.5em}
\caption{Benchmark our \METHOD{} and standard weight finetuning on the test-clean/test-other sets of LibriSpeech on top of the multilingual pretrained xlsr under different low-resource settings.}
\label{fig:ft_xlsr_libri}
\vspace{-1.5em}
\end{figure}

\textbf{Finetuning xlsr on LibriSpeech.} 
We benchmark our \METHOD{} with standard weight finetuning on top of the multilingual SSL pretrained xlsr for low-resource English ASR on LibriSpeech. As shown in Fig.~\ref{fig:ft_xlsr_libri}, our \METHOD{} outperforms standard weight finetuning in terms of the achievable WER, e.g., a 6.02\%/8.33\% reduction in WER on LibriSpeech test-clean/other when being finetuned on 10min labeled data, respectively, indicating the decent scalability and generality of our \METHOD{} even under the large gap between the pretraining resources and finetuning resources.

\begin{wraptable}{r}{0.5\textwidth}
  \vspace{-2em}
  \centering
  \caption{Benchmark our \METHOD{} and weight finetuning on wav2vec2-large and CommonVoice.}
    \resizebox{0.5\textwidth}{!}{
\begin{tabular}{cccccc}
\toprule
\textbf{Language} & \textbf{Dutch} & \textbf{Mandarin} & \textbf{Spanish} & \textbf{Tatar} & \textbf{Russian} \\ \midrule
Weight ft & 19.821 & 26.674 & 13.864 & 11.143 & 17.052 \\
\METHOD{} & \textbf{17.327} & \textbf{25.897} & \textbf{11.887} & \textbf{9.912} & \textbf{15.403} \\ \midrule
\textbf{Language} & \textbf{Italian} & \textbf{Kyrgyz} & \textbf{Turkish} & \textbf{Swedish} & \textbf{France} \\ \midrule
Weight ft & 19.265 & 13.409 & 15.699 & 20.807 & 19.349 \\
\METHOD{} & \textbf{17.425} & \textbf{11.279} & \textbf{13.31} & \textbf{18.664} & \textbf{16.304} \\ \bottomrule
\end{tabular}
    }
  \vspace{-1em}
  \label{ft_wav2vec2_large_cv}%
\end{wraptable}%

\textbf{Finetuning wav2vec2-large on CommonVoice.}
We benchmark our \METHOD{} with standard weight finetuning on top of English pretrained wav2vec2-large for phoneme recognition on other languages from CommonVoice, which is a high-to-low resource transfer setting as a complement to Sec. 4.3 of our manuscript. As shown in Tab.~\ref{ft_wav2vec2_large_cv}, consistent with the results in our manuscript, our \METHOD{} still wins the lowest achievable PER over weight finetuning, which further validates the scalability of our method to larger models under cross-lingual transfer settings.

\begin{figure}[ht!]
  \vspace{-1em}
\centering
\includegraphics[width=0.95\linewidth]{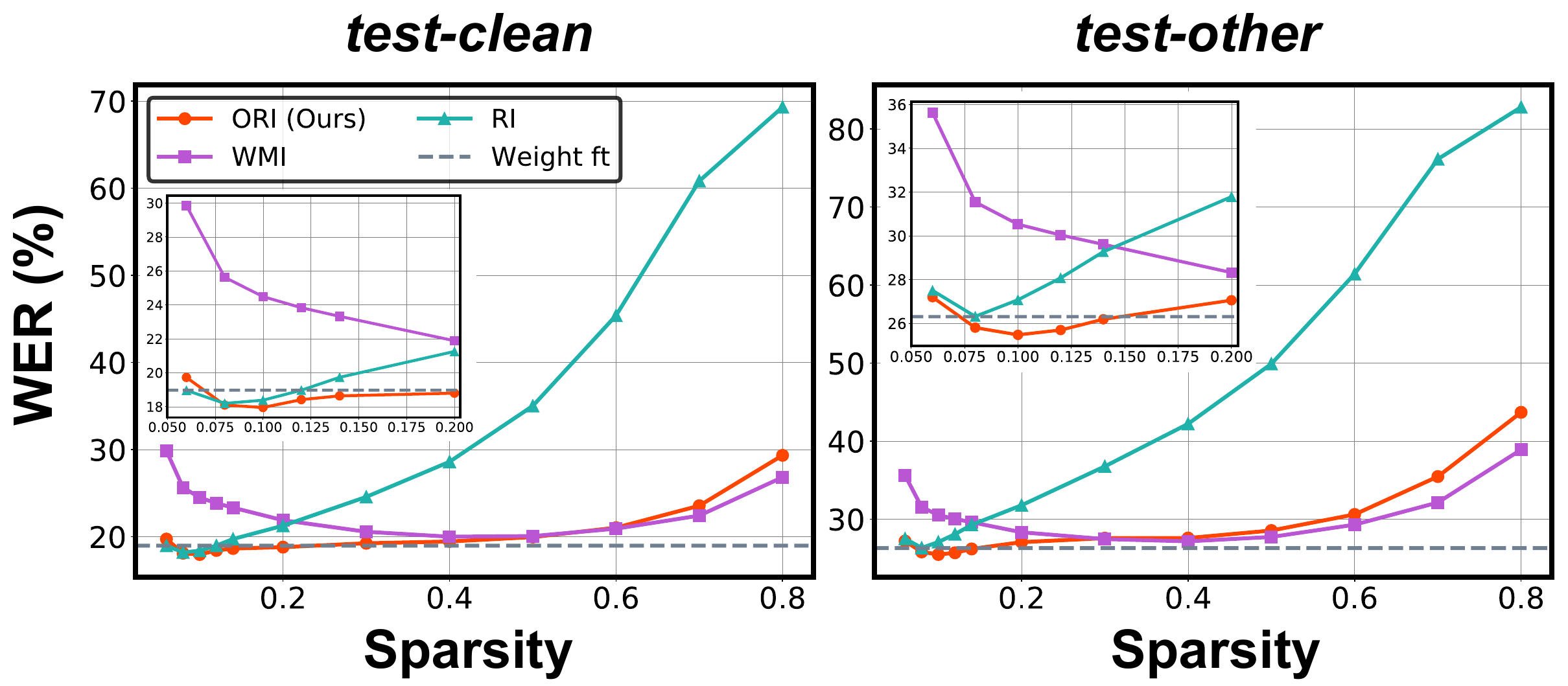}
\vspace{-0.5em}
\caption{Benchmark different mask initialization schemes for finetuning wav2vec2-base on top of LibriSpeech-1h via our \METHOD{}. Both WERs on test-clean (left) and test-other (right) are reported.}
\label{fig:mask_init}
\end{figure}

\textbf{Complete sparsity-WER trade-offs of different mask initialization schemes.}
We provide the complete sparsity-WER trade-offs achieved different mask initialization schemes for finetuning wav2vec2-based on LibriSpeech-1h with our \METHOD{}, as a complement to Tab.1 in Sec. 4.2 of our manuscript, in Fig.~\ref{fig:mask_init}, In addition to the lowest achievable WER of our proposed ORI mask initialization, we can also observe that \underline{(1)} the achievable WER of random mask initialization can match or surpass that of standard weight finetuning under small sparsity ratios while it suffers from a steep increase in WER along with the increase in sparsity ratios due to the lack of utilizing the priors of the pretrained speech representation; \underline{(2)} Weight magnitude based initialization features better scalability to large sparsity ratios, where keeping more important connections, approximately indicated by higher weight magnitudes, becomes more crucial. However, its achievable WER is even inferior to that of random mask initialization due to worse trainability, i.e., it is harder to overturn the ranking between masks via gradients, especially when magnitudes of the pretrained weights cannot serve as a good metric for indicating their importance on downstream speech. Our proposed ORI strategy marries the best of both worlds thus wins the best achievable WER.

\begin{figure}[t!]
\centering
\includegraphics[width=\linewidth]{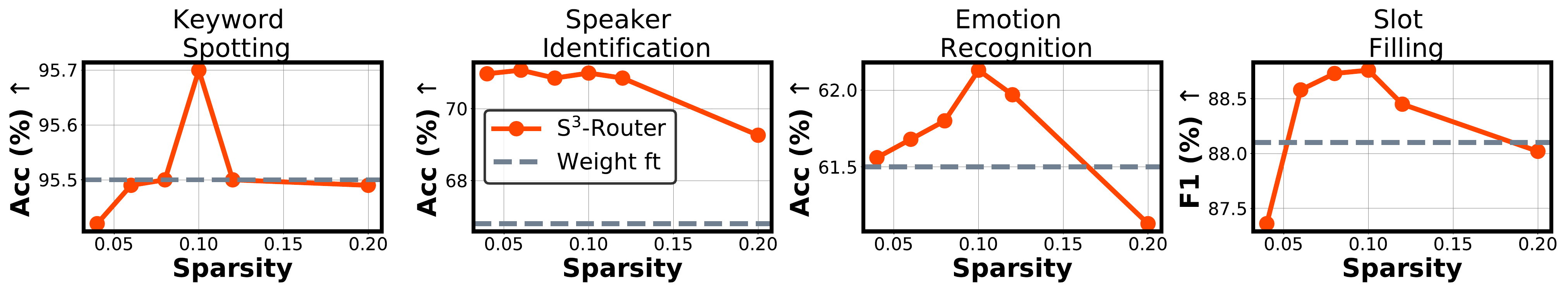}
\vspace{-1.5em}
\caption{Benchmark our \METHOD{} with standard weight finetuning on top of wav2vec2-base on four representative speech processing tasks from SUPERB~\cite{yang2021superb}.}
\label{fig:speech_processing}
\vspace{-0.5em}
\end{figure}

\textbf{Complete sparsity-WER trade-offs on different speech processing tasks.}
We also provide the complete sparsity-WER trade-offs for finetuning wav2vec2-base via our \METHOD{} on different speech processing tasks from SUPERB~\cite{yang2021superb} in Tab. 5/Sec. 4.4 of our manuscript. In particular, we pick one representative task from each of the four task categories for processing different aspects of speech (content/speaker/semantics/paralinguistics). As shown in Fig.~\ref{fig:speech_processing}, we can make a consistent observation as Sec.4 of our manuscript that properly discarding $\leq$ 10\% weights can serve as a decent alternative, featuring better achievable task performances, for weight finetuning.

\section{More Evaluation Results of \METHOD{} as an ASR Pruning Technique}
\label{sec:pruning}

\begin{figure}[ht!]
\centering
\includegraphics[width=\linewidth]{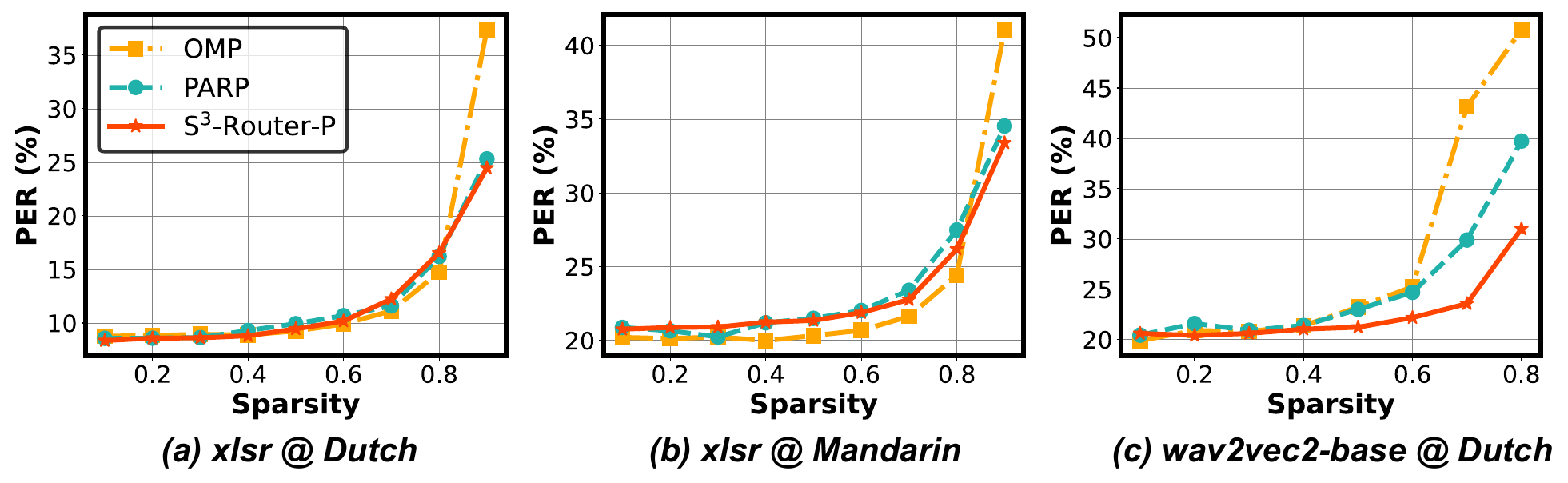}
\vspace{-2em}
\caption{Benchmark our \METHOD{}-P against OMP and PARP for pruning xlsr/wav2vec2-base on Dutch/Mandarin from CommonVoice.}
\label{fig:pr_cv_appendx}
\vspace{-1.5em}
\end{figure}

\textbf{Pruning xlsr/wav2vec2-base on CommonVoice.}
We further evaluate our \METHOD{}-P for ASR pruning on more speech SSL models as a complement to Sec. 5 of our manuscript. As shown in Fig.~\ref{fig:pr_cv_appendx}, our \METHOD{}-P consistently achieves comparable or better pruning performances over PAPR under large sparsity ratios across different models and downstream datasets, which indicates the potential of our method as an all-in-one technique for facilitating the practical usage of speech SSL models.

\begin{wrapfigure}{r}{0.5\textwidth} 
    \centering
    \vspace{-1.5em}
    \includegraphics[width=0.5\textwidth]{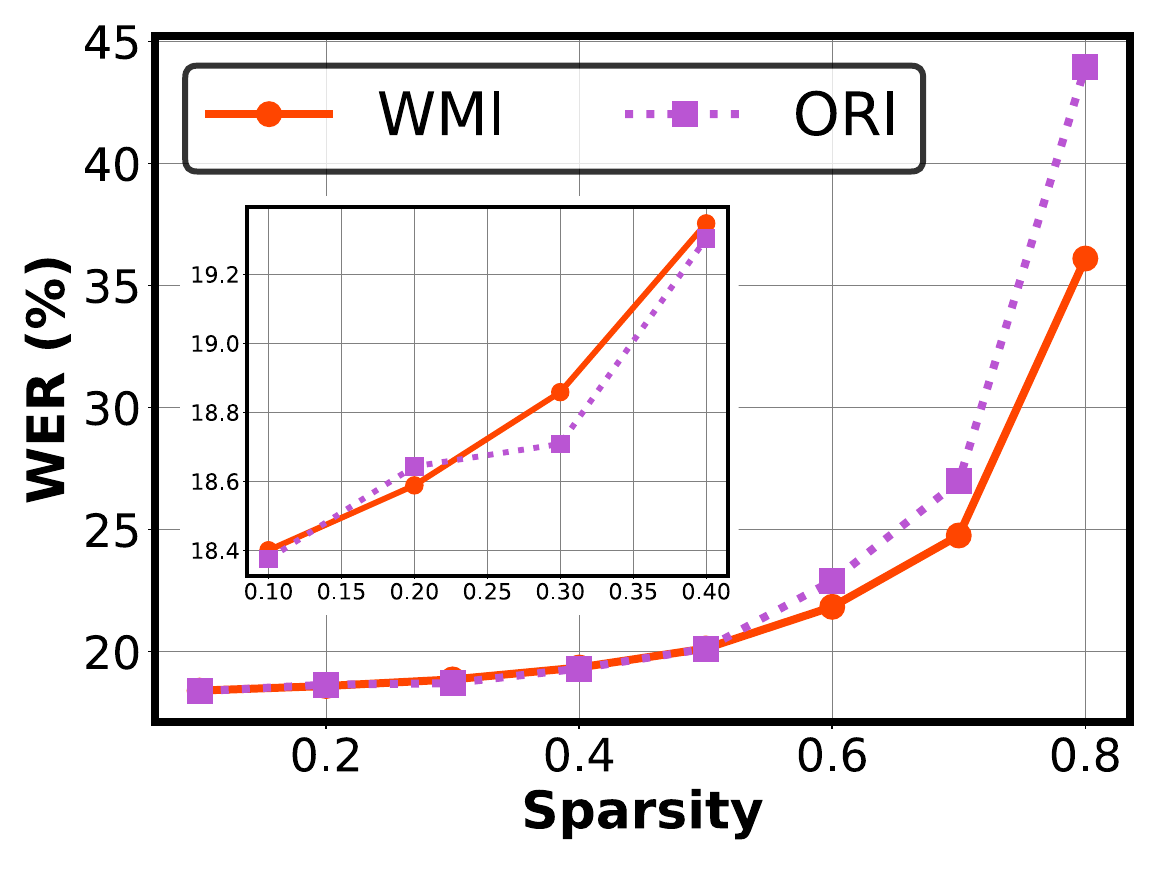}
    \vspace{-2em}
    \caption{Benchmark different mask initialization schemes for pruning wav2vec2-based on top of LibriSpeech-1h. WERs on test-clean are reported.}
    \label{fig:prune_init}
    \vspace{-1.5em}
\end{wrapfigure}

\textbf{Ablation study of the impact of different mask initialization schemes for ASR pruning.}
We adopt different mask initialization schemes in our \METHOD{} for ASR pruning and benchmark their pruning performances in Fig.~\ref{fig:prune_init}. 
We can observe that weight magnitude based mask initialization wins better scalability to large sparsity ratios over ORI, although the latter achieves better finetuning performances as shown in Sec.~\ref{sec:finetuning}. We assume this is because after weight finetuning on downstream speech, the magnitudes of model weights can serve as a better indicator for their importance on the downstream speech, resulting in the better scalability to large sparsity ratios where keeping more important connections becomes more crucial. Therefore, we adopt weight magnitude based mask initialization for ASR pruning by default in our manuscript.

\section{More Analysis about the Properties of Speech SSL Models via \METHOD{}}
\label{sec:correlation}

\begin{figure}[t!]
  \vspace{-1.5em}
\centering
\includegraphics[width=\linewidth]{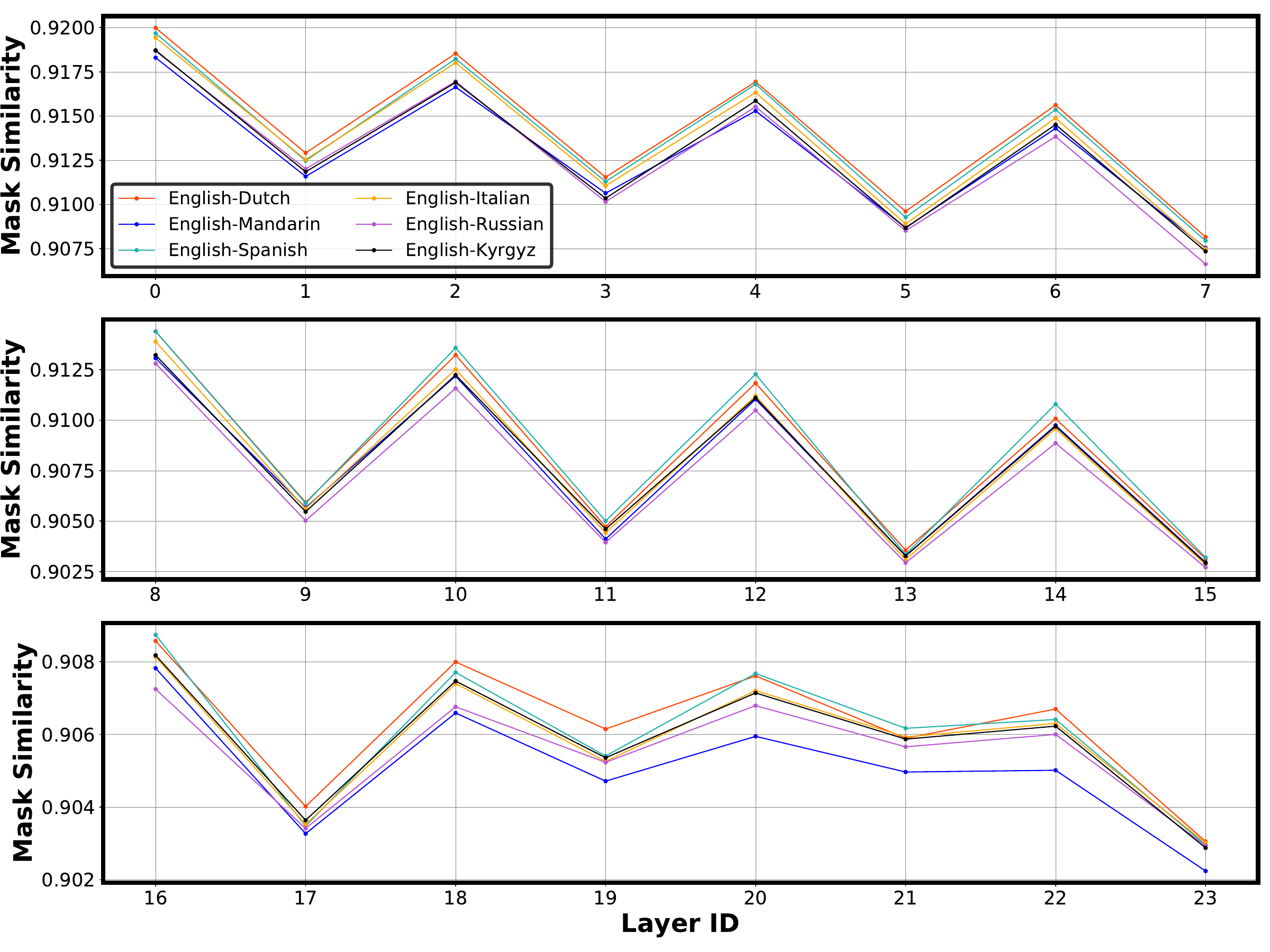}
\vspace{-2em}
\caption{Visualizing the layer-wise cosine similarity of learned masks, on top of the 24 layers in 12 FFNs of wav2vec2-base, between different languages.}
\label{fig:cosine_sim}
\vspace{-1.5em}
\end{figure}

\textbf{Layer-wise cosine similarity of learned masks between different languages.}
We visualize the layer-wise cosine similarity between the learned mask, under a sparsity ratio of 0.1 with near-optimal performances, for English ASR on LibriSpeech-1h and the ones learned on six languages from CommonVoice in Fig.~\ref{fig:cosine_sim}. In particular, there are totally 24 tunable fully-connected layers in the 12 FFNs of wav2vec2-base.
We can observe that \underline{(1)} the learned masks are similar across languages with cosine similarities $\geq$ 0.9; \underline{(2)} Later blocks generally feature lower mask similarities, indicating that more language-specific information are encoded in later blocks; \underline{(3)} The mask similarity ranking keeps relatively stable across consecutive layers; \underline{(4)} The mask similarity generally aligns with human intuitions, e.g., the mask similarity between English and Dutch is higher than that between English and Mandarin, and the quantitative degree of such alignment is indicated by the correlation between their mask similarity and their expert-defined similarity in phonetics.

\begin{wraptable}{r}{0.5\textwidth}
  \vspace{-0.5em}
  \centering
  \caption{Pearson correlation between two similarity metrics calculated on different layers.}
    \resizebox{0.5\textwidth}{!}{
\begin{tabular}{ccccccc}
\toprule
\textbf{Layer ID} & \textbf{0} & \textbf{1} & \textbf{2} & \textbf{3} & \textbf{4} & \textbf{5} \\ \midrule
Pearson Corr & 0.527 & 0.374 & 0.585 & 0.477 & 0.611 & 0.431 \\ \midrule \midrule
\textbf{Layer ID} & \textbf{6} & \textbf{7} & \textbf{8} & \textbf{9} & \textbf{10} & \textbf{11} \\ \midrule
Pearson Corr & 0.561 & 0.372 & 0.539 & 0.366 & 0.436 & 0.351 \\ \midrule \midrule
\textbf{Layer ID} & \textbf{12} & \textbf{13} & \textbf{14} & \textbf{15} & \textbf{16} & \textbf{17} \\ \midrule
Pearson Corr & 0.416 & 0.271 & 0.296 & 0.17 & 0.125 & 0.126 \\ \midrule \midrule
\textbf{Layer ID} & \textbf{18} & \textbf{19} & \textbf{20} & \textbf{21} & \textbf{22} & \textbf{23} \\ \midrule
Pearson Corr & 0.156 & 0.105 & 0.225 & 0.211 & 0.154 & 0.118 \\ \bottomrule
\end{tabular}
    }
  \vspace{-0.5em}
  \label{tab:correlation}%
\end{wraptable}%

\textbf{Correlation between mask similarities and phonetic similarities calculated on different layers.}
As shown in Sec. 6.3 of our manuscript, there exists a non-trivial correlation between the two similarity metrics on the first layer. We further measure the Pearson Correlation Coefficient~\cite{benesty2009pearson} between the phonetic similarity and the mask similarity calculated on top of different layers. 
As shown in Tab.~\ref{tab:correlation}, we can observe non-trivial correlations between the two similarity metrics on early layers while their correlations on later layers are poor. This further indicates that early layers in speech SSL models process  phonetic features extracted from raw audios, which are highly correlated to human expertise in phonetics, while later blocks process more task-specific features, which integrate other high-level information, e.g., for language modeling in ASR.

\bibliographystyle{unsrt}
\bibliography{ref}